\documentclass[10pt,twocolumn,letterpaper]{article}

\usepackage[pagenumbers]{cvpr} % To force page numbers, e.g. for an arXiv version
\usepackage{multirow}
\definecolor{cvprblue}{rgb}{0.21,0.49,0.74}
\usepackage{mathrsfs}
\usepackage{color}
\usepackage{xcolor}
\definecolor{citecolor}{HTML}{0071bc} 
\definecolor{SeaGreen4}{RGB}{0,205,102} 
\definecolor{SlateBlue}{RGB}{106,90,205} 
\definecolor{DarkRed}{RGB}{178,34,34} 
\usepackage[colorlinks, linkcolor=red,  anchorcolor=blue, citecolor=citecolor]{hyperref}

\def\paperID{18363} % *** Enter the Paper ID here
\def\confName{CVPR}
\def\confYear{2025}

\title{ Object Detection using Event Camera: A MoE Heat Conduction based Detector and A New Benchmark Dataset} 

\author{
Xiao Wang$^{1}$, Yu Jin$^{1}$, Wentao Wu$^{2}$, Wei Zhang$^{3}$, Lin Zhu$^{4}$, Bo Jiang$^{1}$, Yonghong Tian$^{3,5,6}$ \\
${^1}$School of Computer Science and Technology, Anhui University, Hefei, China \\
${^2}$School of Artificial Intelligence, Anhui University, Hefei, China \\
${^3}$Peng Cheng Laboratory, Shenzhen, China \\
${^4}$Beijing Institute of Technology, Beijing, China \\
${^5}$National Key Laboratory for Multimedia Information Processing, \\ School of Computer Science, Peking University, China \\
${^6}$School of Electronic and Computer Engineering, Shenzhen Graduate School, Peking University, China
}

\begin{document}
\maketitle

\begin{abstract}
Object detection in event streams has emerged as a cutting-edge research area, demonstrating superior performance in low-light conditions, scenarios with motion blur, and rapid movements. Current detectors leverage spiking neural networks, Transformers, or convolutional neural networks as their core architectures, each with its own set of limitations including restricted performance, high computational overhead, or limited local receptive fields. 
This paper introduces a novel MoE (Mixture of Experts) heat conduction-based object detection algorithm that strikingly balances accuracy and computational efficiency. Initially, we employ a stem network for event data embedding, followed by processing through our innovative MoE-HCO blocks. Each block integrates various expert modules to mimic heat conduction within event streams. Subsequently, an IoU-based query selection module is utilized for efficient token extraction, which is then channeled into a detection head for the final object detection process. 
Furthermore, we are pleased to introduce EvDET200K, a novel benchmark dataset for event-based object detection. Captured with a high-definition Prophesee EVK4-HD event camera, this dataset encompasses 10 distinct categories, 200,000 bounding boxes, and 10,054 samples, each spanning 2 to 5 seconds. 
We also provide comprehensive results from over 15 state-of-the-art detectors, offering a solid foundation for future research and comparison. 
The source code of this paper will be released on: 
\href{https://github.com/Event-AHU/OpenEvDET}{https://github.com/Event-AHU/OpenEvDET}
\end{abstract}
% Additionally, the source code for this paper will be made available on GitHub, enabling the research community to build upon and further refine our methodologies. 
% our detector is flexible, accurate, and efficient for event stream based detection. 

\section{Introduction}~\label{sec1}
Object detection aims to identify predefined target objects by delineating them with bounding boxes and assigning category labels. It stands as a cornerstone problem in computer vision and finds extensive application across fields such as intelligent video surveillance, autonomous vehicles, and industrial automation. With the advent of deep learning, a plethora of cutting-edge deep object detectors have emerged, demonstrating remarkable performance with RGB cameras. Notable examples include the RCNN variants~\cite{girshickICCV15fastrcnn, ren2016fasterrcnnrealtimeobject, he2018maskrcnn}, 
YOLO-based models~\cite{wang2024yolov10, li2022yolov6, luo2024integervaluedtrainingspikedriveninference, wang2024mambayolossmsbasedyolo, bochkovskiy2020yolov4optimalspeedaccuracy, su2023deepdirectlytrainedspikingneural}, and 
DETR-inspired detectors~\cite{zhao2024rtdetr, carion2020endtoendobjectdetectiontransformers, liu2022dabdetrdynamicanchorboxes, zhang2022dinodetrimproveddenoising, zong2023codetrdetrscollaborativehybridassignments, zhu2021deformableDETR}. Nonetheless, frame-based object detectors continue to struggle in demanding conditions such as low light, complex backgrounds, and rapid motion. The constraints in image quality inherent in traditional frame cameras, which capture images at a fixed frame rate (e.g., 30 FPS) and employ a uniform exposure setting, are largely to blame for the prevalence of missed and erroneous detections across various detection algorithms.

\begin{figure*}
\centering
\includegraphics[width=\linewidth]{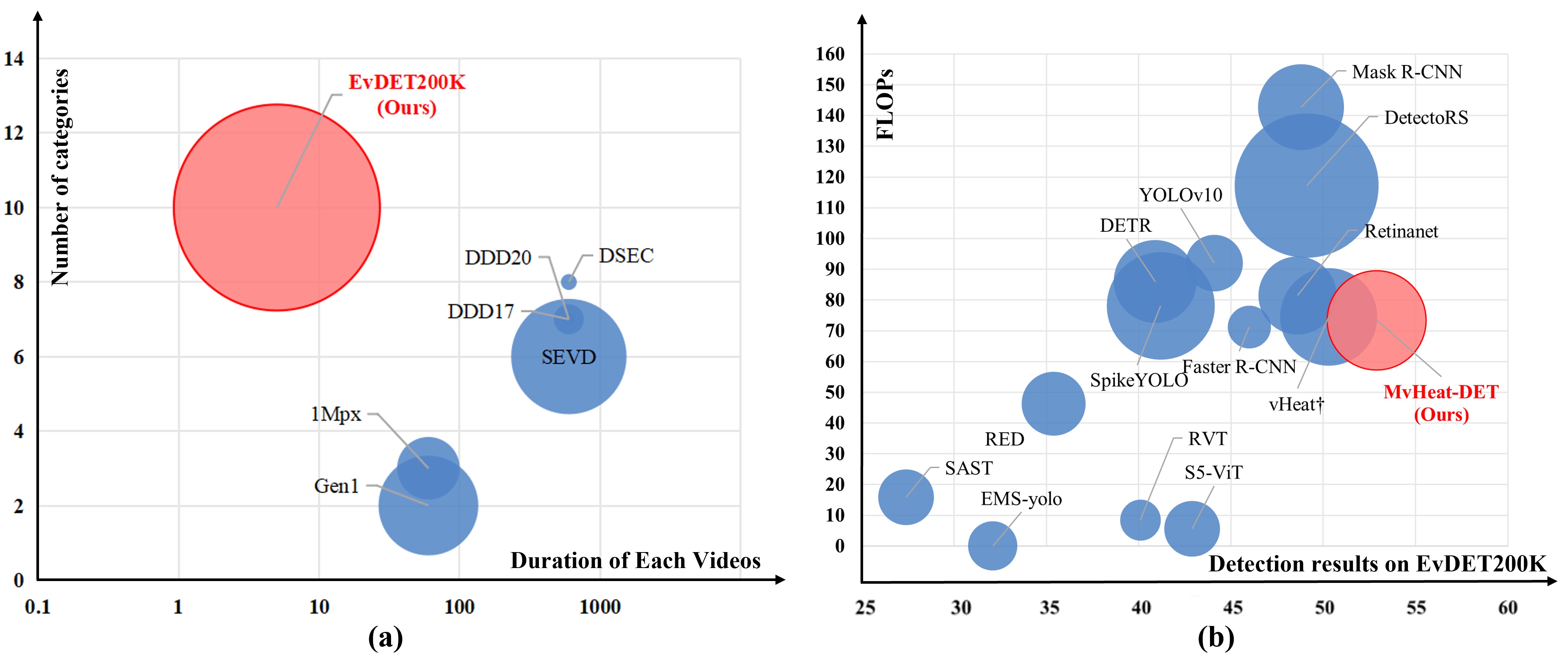}
\caption{ 
(a). Comparison of existing datasets and our proposed EvDET200K dataset for event stream based detection; 
(b). Comparison of our proposed MvHeat-DET and existing SOTA detectors on the EvDET200K dataset. 
}
\label{fig_dataset_benchmark_bubble}
\end{figure*}

To overcome the limitations of traditional sensors, researchers have turned to innovative technologies for object detection. Among these, bio-inspired event cameras, also known as Dynamic Vision Sensors (DVS), have garnered significant interest within the computer vision sphere. Event cameras outshine conventional RGB frame-based cameras in several aspects: \textit{high dynamic range}, \textit{high temporal resolution}, \textit{low energy consumption}, and nearly no motion blur. These sensors have been applied to a range of tasks, from high-level applications such as event-based tracking~\cite{wang2024eventvot}, recognition~\cite{wang2024hardvs}, and captioning~\cite{zhang2025evsign, wang2024eventSign}; to low-level operations including image enhancement and reconstruction~\cite{zhu2022ultrareconstruction}. 
In the realm of event-based object detection, DAGr~\cite{gehrig2024nature}, introduced by Daniel et al., integrates RGB frames with event streams to strike an optimal balance between latency and bandwidth, all while maintaining detection precision. SpikeYOLO~\cite{luo2024integervaluedtrainingspikedriveninference} has made strides by incorporating the I-LIF spiking neuron, along with integer training and spike-driven inference, to minimize quantization errors in Spiking Neural Networks (SNNs). 

% This advancement not only enhances detection accuracy but also preserves the low power consumption attribute of event cameras.

% RT-DETR~\cite{zhao2024rtdetr} design an efficient hybrid encoder and query selection strategy, effectively addressing issues such as slow training and NMS latency in real-time object detection. 

% Although significant breakthroughs have been made, there are still the following issues with current event-based object detectors: 
% 1). Current event-based detectors usually adopt the CNN (Convolutional Neural Networks), Transformer as their backbone network, however, the performance of CNNs is limited by local receptive fields and fails to capture long-range and complex dependencies. Meanwhile, the Transformer-based vision model, such as ViT~\cite{dosovitskiy2021vit}, suffers from relatively high computational complexity $\mathcal{O}(N^2)$ and a lack of interpretability. 
% 2). Some researchers adopt the bio-inspired SNN (Spiking Neural Networks)~\cite{su2023deepdirectlytrainedspikingneural, luo2024integervaluedtrainingspikedriveninference} to encode the event streams and gain an advantage in energy consumption, but the overall performance is significantly lower than that of the ANN (Artificial Neural Networks)-based detector. 
% %% 
% Therefore, it is still a challenging task to design an effective, efficient, and interpretable event-based object detection algorithm. 

Despite the notable advancements made, current event-based object detectors still face the following issues: 
\textbf{1).} Most event-based detectors rely on CNNs (Convolutional Neural Networks) or Transformers as their backbone architectures. However, CNNs are constrained by their local receptive fields, which hinders their ability to capture long-range and intricate dependencies. In contrast, Transformer-based vision models, such as ViT~\cite{dosovitskiy2021vit}, grapple with high computational complexity, scaling as $\mathcal{O}(N^2)$, and lack interpretability. 
\textbf{2).} Some researchers resort to bio-inspired SNNs (Spiking Neural Networks)~\cite{su2023deepdirectlytrainedspikingneural, luo2024integervaluedtrainingspikedriveninference} for encoding event streams, reaping benefits in terms of energy efficiency. Nonetheless, their overall performance lags significantly behind that of ANN (Artificial Neural Network)-based detectors.
Consequently, the quest for an effective, efficient, and interpretable event-based object detection algorithm remains a formidable challenge.

Recently, Wang et al. proposed a physics-inspired vision backbone model, vHeat~\cite{wang2024vheatbuildingvisionmodels}, which is grounded in the principles of heat conduction. The core module of this model is the Heat Conduction Operator (HCO), which envisions image patches as heat sources and conceptualizes the determination of their correlations as the process of thermal energy diffusion. By employing 2D Discrete Cosine Transform (DCT) and Inverse Discrete Cosine Transform (IDCT) operations to approximate the HCO, they achieve a lower computational complexity of $\mathcal{O}(N^{1.5})$. When applied to object detection, the HCO outperforms both the Swin-Transformer~\cite{liu2021swintransformerhierarchicalvision} and ConvNeXt~\cite{liu2022convnet2020s}. However, we argue that the 2D DCT and IDCT operators may not be the most suitable for simulating heat conduction in the context of vision models. As these operators are designed for general signal processing, they may not fully capture the spatial and temporal dynamics inherent in visual data, which are critical for accurate object detection. This motivates us to explore alternative approaches that can more accurately and efficiently model the heat conduction process for event-based detection.

In this paper, we propose a novel backbone network for event-based detection that consists of Mixture-of-Expert (MoE) based heat conduction operators, termed MvHeat-DET. We split the input event streams into multiple clips and get their feature embeddings using a stem network. Then, we pass the event embeddings into an MHCO layer, as shown in Fig.~\ref{fig_framework}, which first selects a transform branch using the policy network. Then, randomly initialized features FEs are fed into a linear layer to predict the thermal diffusivity and multiplied by the transformed representations. After that, an inverse operator is adopted for complete signal transformation. The obtained tokens will be fed into an IoU-based query selection module and detection head for object localization and recognition by following~\cite{zhao2024rtdetr}.

Although several event stream object detection datasets have been proposed, e.g., Gen1~\cite{detournemire2020gen1largescaleeventbaseddetection} and 1Mpx~\cite{perot20201mpxlearningdetectobjects1}, they are still relatively scarce compared to RGB frame based detection datasets. Therefore, this paper introduces a new benchmark dataset to fill this gap, named EvDET200K. It involves ten categories of target objects, including \textit{people, cars, bicycles, electric bicycles, basketball, ping pong, goose, cats, birds,} and \textit{UAVs}. The dataset was captured using the Prophesee EVK4-HD event camera and comprises 10,054 samples, each ranging from 2 to 5 seconds in duration. An extensive annotation effort has yielded 200,000 high-quality bounding boxes, and we provide over 15 state-of-the-art detectors for future research to benchmark against. We are confident that the introduction of EvDET200K, along with the associated benchmark algorithms, will mark a significant stride forward in the realm of event camera-based object detection.

To sum up, we draw the contributions of this paper as the following three aspects: 

1). We introduce a novel Mixture-of-Experts (MoE)-based heat conduction framework, named MvHeat-DET, designed for event stream object detection. This framework strikingly balances performance, efficiency, and interpretability, offering an improved trade-off in these critical areas.

2). We present EvDET200K, a new high-definition benchmark dataset for event stream object detection. Comprising 10,054 samples captured with the Prophesee EVK4-HD camera, each sample spans 2 to 5 seconds and includes 200,000 bounding boxes spanning 10 distinct object categories. 

3). We have re-trained and evaluated over 15 state-of-the-art (SOTA) object detectors, including models from the YOLO, RCNN, and DETR families, on the newly introduced EvDET200K dataset. This provides a comprehensive baseline for future research to compare and build upon.

\section{Related Works}

\noindent $\bullet$ \textbf{RGB Frame based Detection.~} 
Object detection using RGB images has progressed significantly in recent years, primarily due to advancements in deep learning. These detectors can be divided into three main streams, i.e., RCNN-based detectors~\cite{girshickICCV15fastrcnn, ren2016fasterrcnnrealtimeobject, he2018maskrcnn, attentionrcnn}, YOLO series, and DETR-based detection algorithms. 
% More in detail, RCNN-based detection algorithms typically generate candidate regions, and then extract features for each candidate region. 
% Fast R-CNN~\cite{girshickICCV15fastrcnn} introduced a region pooling layer that performs feature pooling for each candidate region, significantly improving the speed of the algorithm. 
% Faster R-CNN~\cite{ren2016fasterrcnnrealtimeobject} proposed the Region Proposal Network (RPN), enabling the region proposal generation and object detection to be trained in an end-to-end manner. 
% Mask R-CNN~\cite{he2018maskrcnn} further extended Faster R-CNN by adding instance segmentation, providing more fine-grained object information, though at the cost of increased computational and storage overhead. 
% Subsequently, RCNN-based methods have continuously evolved, giving rise to many classic and efficient detection algorithms, such as R-FCN~\cite{dai2023rfcn}, Attention R-CNN~\cite{attentionrcnn}, Pyramid R-CNN~\cite{mao2021pyramidrcnn}, and Sparse R-CNN~\cite{sun2021sparsercnn}. Notably, FPN~\cite{lin2017fpn} has provided an effective solution for handling multi-scale information and has been widely adopted in various detection algorithms.
The DETR~\cite{carion2020endtoendobjectdetectiontransformers} model simplifies the object detection process, effectively eliminating the need for many manually designed components, such as non-maximum suppression (NMS) or anchor generation. 
Based on DETR, methods like Deformable DETR~\cite{zhu2021deformableDETR}, Adaptive Clustering Transformer~\cite{zheng2021adaptiveclusteringtransformer}, PnP-DETR~\cite{wang2022pnpdetr}, and Sparse DETR~\cite{roh2022sparsedetr} reduce computational resource consumption by applying sparse processing to the transformer’s attention mechanism, achieving a favorable balance between efficiency and accuracy. 
% DN-DETR~\cite{li2022dndetr} attributes DETR's slow training convergence to instability in bipartite matching and addresses this by training with noise-augmented ground truths to reconstruct bounding boxes, aiding effective model convergence. 
% DINO~\cite{zhang2022dinodetrimproveddenoising} enhances prior DETR models in performance and efficiency by employing a contrastive denoising training approach and a Mixed Query Selection strategy. 
Furthermore, models like Conditional DETR~\cite{meng2023conditionaldetr}, Anchor DETR~\cite{wang2022anchordetr}, Efficient DETR~\cite{yao2021efficientdetr}, and Dynamic DETR~\cite{DynamicDETR} leverage spatial prior knowledge to better focus on regions of interest (ROIs), significantly reducing learning difficulty. Notably, ~\cite{sun2021rethinkingtransformerbasedsetprediction} proposes integrating R-CNN with Transformer to promote DETR convergence and yield better results.
Different from these Transformer-based detectors, in this paper, we propose a novel MoE Heat Conduction based backbone network for event stream based detection algorithm.

\noindent $\bullet$ \textbf{Event Stream based Detection.~}  
Event-based object detection has recently gained traction, particularly with the development of neuromorphic sensors that operate in an asynchronous, event-driven manner. 
% Unlike traditional frame-based methods that process fixed-image intervals, event-based approaches focus on real-time changes in the scene, capturing pixel-level movements and variations in brightness. 
Due to the specific features of event streams, current researchers usually adopt SNN (Spiking Neural Networks), Graph Neural Networks (GNN), LSTM (Long-short Term Memory) for event-based object detection. To be specific, in methods based on SNNs, SpikingYOLO~\cite{kim2019spikingyolo}, SpikeYOLO~\cite{luo2024integervaluedtrainingspikedriveninference}, and EMS-YOLO~\cite{su2023deepdirectlytrainedspikingneural} combine YOLO with SNNs. SpikingYOLO is the first object detection model implemented in deep SNNs. 
SpikeYOLO introduces the I-LIF spiking neuron to reduce quantization errors in SNNs. 
% EMS-YOLO proposes a novel directly trained spiking neural network for object detection. 
SFOD~\cite{fan2024sfodspikingfusionobject} achieves multi-scale feature map fusion in SNNs for the first time, improving the model's ability to detect objects of various sizes.
% ~\cite{cordone2022objectdetectionspikingneural} encodes event data into voxel cubes, preserving the binary and temporal information of the events while reducing the number of time steps, and facilitating processing by SNNs.
% AEGNN~\cite{schaefer2022aegnnasynchronouseventbasedgraph} models event data as a spatiotemporal graph and uses graph neural networks to process it, reducing the computational burden of event data. 
DAG-r~\cite{gehrig2022eagr} employs an efficient asynchronous graph neural network to handle event data, achieving a trade-off between bandwidth and latency. 
% EAGR~\cite{gehrig2022eagr} treats events as spatiotemporal evolving event graphs and can be deployed in an efficient, asynchronous mode. 
% DMANet~\cite{wang2023dmanet} dynamically aggregates spatiotemporal information in event streams using Long-Short Term Memory modules (LRM and SRM)~\cite{vennerød2021lstm}, improving target localization and recognition.
% RVT~\cite{gehrig2023recurrentvisiontransformersobject} uses LSTM~\cite{vennerød2021lstm} units for temporal feature aggregation, preserving temporal information while minimizing latency. 
Additionally, AED~\cite{Liu2023aed} is a lightweight detector with fast detection speed, better suited to the high temporal resolution of event cameras.
GET~\cite{peng2023Get} introduces the EDSA block, which effectively extracts features and enables feature communication in the spatial and time-polarity domains.
S5-ViT~\cite{zubić2024statespacemodelsevent} performs temporal aggregation using a state-space model (SSM), addressing the challenge of RNNs' limited generalization when handling inputs with varying frequencies.

% For example, SpikeYOLO~\cite{luo2024integervaluedtrainingspikedriveninference} adapts the YOLO framework for event-based settings using spiking neural networks (SNNs). 
% This integration optimizes performance in energy-constrained environments and demonstrates the potential of event-driven models for real-time detection. 
% Additionally, GWD~\cite{zubić2023chaoscomesorderordering} and GET~\cite{peng2023Get} explore geometric and graph-enhanced techniques to improve robustness and contextual understanding. 
% While these approaches mainly target frame-based detection, their principles can also enhance event-based detection by leveraging the unique properties of event data. Furthermore, incorporating temporal dynamics, as seen in S5-ViT~\cite{zubić2024statespacemodelsevent}, underscores the importance of understanding object movements over time. This trend highlights the need for innovative architectures that effectively process event-based data, paving the way for more efficient real-time applications. 

\noindent $\bullet$ \textbf{Benchmark Datasets for Event-based Detection.~}  
Event-based vision has gained significant attention due to its high temporal resolution and its ability to handle challenging conditions such as fast motion and varying lighting. To advance object detection in this domain, several notable datasets have been proposed. The SEVD~\cite{aliminati2024sevdsyntheticeventbasedvision} dataset provides a comprehensive synthetic event-based dataset for both ego-centric and fixed-camera traffic perception, allowing researchers to explore complex traffic monitoring scenarios. Similarly, the eTraM~\cite{verma2024etrameventbasedtrafficmonitoring} dataset captures real-world traffic scenes using neuromorphic sensors, specifically designed for vehicle detection and tracking in urban environments. In the realm of automotive applications, the Gen1~\cite{detournemire2020gen1largescaleeventbaseddetection} dataset introduces high-resolution event data recorded from vehicles, enabling precise object detection in high-speed situations and under challenging lighting conditions. Extending these capabilities further, the 1Mpx~\cite{perot20201mpxlearningdetectobjects1} dataset offers even finer detail for object detection, particularly useful in dynamic and low-light environments. 
% To make traditional image datasets more applicable to event-based processing, the N-Caltech101~\cite{orchard2015ncaltechconvertingstaticimagedatasets} dataset proposes a method to convert conventional image datasets into spiking neuromorphic data, enhancing the usability of existing resources. 
% For safety-critical applications such as healthcare, the Pedestrian Detection Dataset~\cite{miao2019PedestrianDetectionDataset} provides event-based recordings tailored to pedestrian monitoring, behavior analysis, and fall detection tasks. 
% For real-time object detection, the DDD17~\cite{binas2017ddd17endtoenddavisdriving} dataset offers synchronized event and frame data from a DAVIS camera, facilitating object detection and scene understanding in dynamic driving environments. 
These datasets mark significant advancements in event-based object detection, offering diverse benchmarks that span various domains. 
%%%% 
Different from these datasets, our proposed EvDET200K dataset provides high-definition event streams captured under different weathers and lightings and involves 10 categories.

\section{MvHeat-DET}

\subsection{Preliminaries: Physical Heat Conduction} 
In an ideal scenario, the temperature at a point with coordinates \( (x, y) \) in a two-dimensional object at time \( t \), marked  \( u(x, y, t) \), is governed by the following heat conduction equation in an isotropic medium:
\begin{equation}
\small 
\label{eq1}
\frac{\partial u(x, y, t)}{\partial t} = k \left( \frac{\partial^2 u(x, y, t)}{\partial x^2} + \frac{\partial^2 u(x, y, t)}{\partial y^2} \right) = k(u_{xx} + u_{yy})
\end{equation}
where \( k \) is the thermal diffusivity, which measures the efficiency of heat diffusion within the material. To find the general solution of Eq.~\ref{eq1}, we apply the Fourier transform to both sides of the equation, rewriting it as:
\begin{equation}
\small 
\label{eq2}
\mathcal{F} \frac{ \partial{u}(x, y, t)}{\partial t} = k\mathcal{F} (u_{xx} + u_{yy})
\end{equation}
where \( \mathcal{F} \) is the Fourier transform function. Let \( \hat{u}(v_x, v_y, t) \) denotes the Fourier transform of \( u(x, y, t) \), and \( v_x \) and \( v_y \) are the frequency variables in Fourier space.  
This transformation converts the partial differential equation into an algebraic equation, which is simpler to solve. Therefore, we can rewrite the Eq.~\ref{eq2} as:
\begin{equation}
\label{eq3}
\frac{\partial \hat{u}(v_x, v_y, t)}{\partial t} = -k (v_x^2 + v_y^2) \hat{u}(v_x, v_y, t)
\end{equation}
where \( t = 0 \) represent the initial state of the object, i.e., \( u(x, y, t)\mid_{t=0} \). For short, we use \( f(x, y) \) instead, and \( \hat f(v_x, v_y) \) denotes the FT-transformed \( f(x, y) \). By setting the initial state in Eq.~\ref{eq3}, we can get the following solution: 
\begin{equation}
\label{eq4}
 \hat{u}(v_x, v_y, t) = \hat f(v_x, v_y)e^{-k(v_x^2+v_y^2)t}
\end{equation}

To obtain a general solution of the heat equation in the spatial domain, we apply the inverse Fourier transform (denoted as \( \mathcal{F}^{-1} \)) to Eq.~\ref{eq4} and get the following expression:
\begin{equation}
\label{eq_final}
u(x, y, t) = \mathcal{F}^{-1}( \hat f(v_x, v_y)e^{-k(v_x^2+v_y^2)t}) 
\end{equation}

Inspired by the aforementioned process, Wang et al. propose a new vision backbone vHeat~\cite{wang2024vheatbuildingvisionmodels} which is built based on Heat Conduction Operator (HCO). They adopt the 2D discrete cosine transformation $DCT_{2D}$ and the 2D inverse discrete cosine transformation $IDCT_{2D}$ to simulate the HCO process in the visual domain. Despite good results that can be obtained, we think it can be further extended as the DCT-IDCT transformation may not be optimal for such a simulation. In the following subsections, we will introduce our MoE heat conduction based backbone network for event-based object detection.

\subsection{Overview} 

\begin{figure*}
\centering
\includegraphics[width=0.85\linewidth]{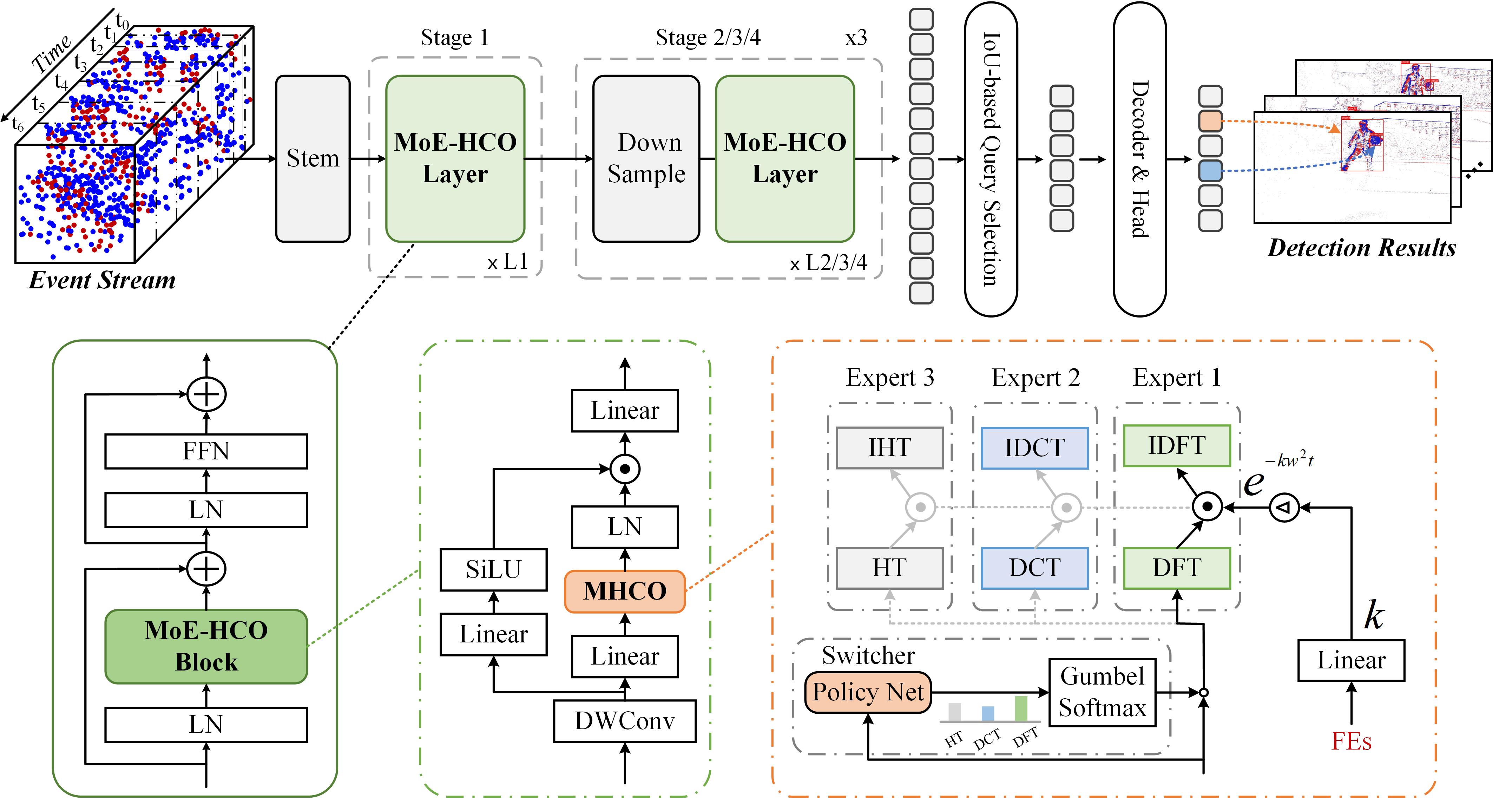}
\caption{An overview of our proposed event-based object detection framework, termed MvHeat-DET.} 
\label{fig_framework}
\end{figure*}

As shown in Fig.~\ref{fig_framework}, given the event streams, we first stack into event frames and get the event embeddings using a stem network. Then, we feed the event embeddings into the MoE-HCO blocks which provides multiple transform candidates. In this work, we consider DFT-IDFT, DCT-IDCT, and HT-IHT as three experts for the validation. A policy network with Gumbel Softmax is utilized for expert selection. In addition, the Frequency Embeddings (FEs) are used to predict the thermal diffusivity and multiplied by the transformed frequency representations. More importantly, we adopt an IoU-based query selection module to find the key tokens for final detection.

% Event cameras possess advantages such as low latency, low power consumption, and high dynamic range, providing significant benefits in certain vision tasks. In this study, we attempt to perform object detection using only event data. 
% To achieve high accuracy while maintaining low time complexity in our network architecture, we introduce the MvHeat model as a substitute for the traditional Transformer encoder. The MvHeat model, built on the principles of thermal energy diffusion, offers a time complexity of \( O(n^{1.5})\), compared to conventional Transformers. Building on this, we implement a mixture-of-experts mechanism to process different input data uniquely, maximizing the potential of specific features within each data space. Additionally, we employ an uncertainty-based token selection module to filter out less relevant tokens before passing the results into the Transformer decoder and downstream task head, yielding the final prediction.

\subsection{MvHeat Backbone Network}  

Drawing inspiration from the physical concept of thermal conduction, we investigate the spread of image features across spatial domains through the lens of heat diffusion, culminating in the development of the MvHeat backbone network. This network harnesses the MoE (Mixture of Experts) Heat Conduction Operation, a novel framework that adapts thermal conduction principles for the processing of discrete visual data features. The MvHeat backbone is engineered to deliver specialized processing for unique image features, thereby enhancing their integration and analysis. 

\noindent $\bullet$ \textbf{MoE-HCO Block.~}  
The MvHeat Encoder is structured into four stages, each comprising \( L_i, i = \{1, 2, 3, 4\} \) MHCO (MoE Heat Conduction Operation) Layers. As the data progresses through each stage, the spatial resolution is reduced by half via downsampling, followed by processing through multiple MHCO layers. The MHCO architecture is closely similar to Vision Transformer (ViT), with a pivotal distinction: it substitutes ViT's attention blocks with our innovative method while retaining the rest of the architectural framework. The efficacy of this structure has been validated by prior Transformer-based research, ensuring both scalability and a reduction in the computational burden associated with traditional attention mechanisms. In the MHCO module, input data is initially funneled through a selection mechanism to identify the optimal expert branch for the current feature set. Subsequently, the thermal diffusivity \(k\) is learned through Feature Embeddings (FEs) and multiplied by a coefficient matrix \(e^{-kw^2t} \) to generate an intermediate output. This intermediate result is then multiplied by the input data transformed into the frequency domain before being reconverted to the temporal domain.

\noindent $\bullet$ \textbf{MHCO: MoE Heat Conduction Operator.~}  
As described in the section on Physical Heat Conduction, we design the MoE Heat Conduction Operator (MHCO) module to extract visual features by fully simulating thermal diffusion. This module enables effective visual feature extraction and facilitates the exchange of visual information between different image patches. Specifically, we first use deep convolution to extend the temperature distribution in the two-dimensional space along the channel dimension, with the resulting multi-channel features denoted as \( U_0 \). Then, to obtain the output \( U_t \) after thermal diffusion, we apply Eq~\ref{eq_final}:
\begin{equation}
\label{eq6}
 U_t = \mathcal{F}^{-1}( \mathcal{F}(U_0)e^{-k(v_x^2+v_y^2)t}) 
\end{equation}
Following a logical reasoning process, we utilize the Discrete Fourier Transform (DFT) to transform discrete image patch features into the frequency domain and then apply the inverse Fourier Transform (IDFT) to revert them back to the spatial domain. In the physical sciences, when considering that a medium does not occupy the entire space, a unique solution to the equation requires specifying boundary conditions for \( u \) as indicated in Eq.~\ref{eq_final}. Additionally, each image patch can be viewed as a diffusion of features within a bounded space. Given that visual data is spatially constrained and semantic information does not propagate beyond the boundaries, a natural boundary condition arises. Here, we introduce a common Neumann boundary condition \( D \):
\begin{equation}
\label{eq7}
 \frac{\partial {u(x,y,t)}}{ \partial \textbf{n}} = 0, \forall(x,y) \in D, t>0
\end{equation}
where \( \textbf{n} \)  denotes the normal to the image boundary D.
Additionally, because visual data is typically rectangular, this boundary condition enables us to perform reasonable transformations using the 2D Discrete Cosine Transform (DCT), 2D inverse Discrete Cosine Transform (IDCT), and Haar Transform (HT), inverse Haar Transform (IHT). Thus, Eq.~\ref{eq6} can be also rewritten as: 
\begin{equation}
\label{eq8}
 U_t = \mathcal{C}^{-1}( \mathcal{C}(U_0)e^{-k(v_x^2+v_y^2)t}) 
\end{equation}
\begin{equation}
\label{eq9}
 U_t = \mathcal{H}^{-1}( \mathcal{H}(U_0)e^{-k(v_x^2+v_y^2)t}) 
\end{equation}
where \( \mathcal{C} \) denotes DCT, \(\mathcal{C}^{-1}\) denotes IDCT and \( \mathcal{H} \) denotes HT, \(\mathcal{H}^{-1}\) denotes IHT. 
We attempt to construct an expert network utilizing three methods simultaneously. This approach is motivated by the characteristics of event data: when capturing scenes with low-motion objects or sparse areas, there tend to be large blank regions. In such cases, after segmenting the original image into patches, the area of interest may only appear within a single patch. Under these circumstances, it is sufficient to limit the network's focus to an individual patch, disregarding interactions between patches. For this, we can apply DCT and HT transformations within patch boundaries alone.
Conversely, when there are high-motion objects or dense scenes, it becomes necessary to account for interactions between patches. In these situations, we leverage DFT from the original inference process to enable effective inter-patch information exchange. As shown in Fig.~\ref{fig_framework}, we input multi-channel data into a selection branch, where a lightweight policy network computes the scores for the current data across three branches. We employ the Gumbel-Softmax trick to ensure that the decision-making process is differentiable.

In physical heat conduction, the thermal diffusivity \(k\) indicates how quickly heat spreads within a material. In visual heat conduction, we assume that the most noteworthy content in the image carries more "heat", and thus, visual heat should flow toward these regions. Naturally, the thermal diffusivity parameter \(k\) should be learnable and adapt to the image content, enhancing the adaptability of heat diffusion to the learning of visual representations.After applying the DCT, DFT or HT transformation, the input data \(x\) is converted into the frequency domain (denoted as \( \hat{x} \) ). Therefore, the learnable thermal diffusivity \(k\) also needs to be derived from the frequency domain information. Inspired by the positional embeddings in ViT, we randomly initialize a Frequency Embeddings (FEs) with the same shape as \( \hat{x} \), which is then fed into a linear layer to predict the thermal diffusivity \(k\). Specifically, we set a fixed value for \( t \), and the FEs is utilized across each stage of the MvHeat network to enhance convergence throughout the training process.

\subsection{IQS: IoU-based Query Selection} 
In the DETR model, object queries are a set of learnable embeddings. To reduce the difficulty of optimizing object queries, some studies have proposed query selection schemes, which typically leverage classification scores to select the top K features from the encoder. However, due to discrepancies between the distributions of classification scores and localization confidence, some predicted boxes with high classification scores may not be close to the ground truth. This results in a bias toward selecting boxes with high classification scores but low IoU scores, while overlooking boxes with lower classification scores but higher IoU scores. Such selection biases undermine the overall performance of the detector.
During training, the IoU score is incorporated into the objective function of the classification branch, encouraging the model to associate high classification scores with ground-truth boxes that have high IoU scores. Thus, we can still select the top K based on classification scores to obtain higher-quality queries. The overall loss can be expressed as follows:
\begin{equation}
\label{eq10}
 \mathcal{L}(y, \hat{y}) = \mathcal{L}_{bbox}(b,\hat{b})+\mathcal{L}_{cls}(IoU,c,\hat{c}) 
\end{equation}
where $y = \{b, c\}$ denotes the ground truth boxes and categories, and $\hat{y} = \{\hat{b}, \hat{c}\}$ represents the predicted boxes and categories.

% \subsection{Training and Inference}  

\section{EvDET200K Benchmark Dataset}

% The dataset comprises event-based data captured across various scenarios, including both indoor and outdoor environments such as schools, streets, and parks. The data collection methods are diverse, featuring handheld, tripod-mounted, and vehicle-mounted perspectives. To ensure a rich variety of data within the same scene, we employed multiple angles for capturing objects. For instance, in vehicle-mounted scenarios, objects in motion were recorded from three distinct directions: front, rear, and side views. This approach enhances the dataset's robustness, making it suitable for tasks requiring diverse and comprehensive event representations.

\subsection{Protocols} 
We aim to provide a good platform for the training and evaluation of event-based object detection. When constructing the EvDET200K benchmark dataset, we follow the following protocols:
\textbf{1). Large-scale:} With the deep integration of information technology, human production and life, large-scale datasets show an increasingly important position. In our work, we collect more than 10k event sequences, totally about 200k objects from 10 classes. 
\textbf{2). Diversity:} During the shooting process, we anticipated potential challenges and configured certain factors in advance. These factors may affect the performance of the data captured by the sensor in object detection tasks. More in detail, Multi-view, Multi-illumination, Multi-motion, Dynamic Background, Non-detection Interference are all considered when recording these event streams. 
\textbf{3). Small Object:} We focus on enhancing the detection capability for small objects. Since small objects are often overlooked in detection tasks, we specifically plan to capture data from multiple perspectives to ensure diversity across different scenarios. Finally, the dataset contains 51\% small objects, providing a sufficient number of samples for training.

\begin{table*}[!htp]
\renewcommand\arraystretch{1.2}
    \centering
    \caption{Comparison of event datasets for object detection. (CL: clear, RA: rainy, DT: daytime, NT: nighttime, MS: multi-scene, MM: multi-motion.)} 
    \resizebox{\textwidth}{!}{ % 这里缩放表格以适应页面宽度
        \begin{tabular}{c|c|l|c|c|c|c|c|c|cc|cc|ccc}
            \hline 
            \multirow{2}{*}{\textbf{Dataset}} & \multirow{2}{*}{\textbf{Year}} & \multirow{2}{*}{\textbf{Sensor}} & \multirow{2}{*}{\textbf{Resolution}} 
            & \multirow{2}{*}{\textbf{Scale}} & \multirow{2}{*}{\textbf{Bbox}} & \multirow{2}{*}{\textbf{Duration}} & \multirow{2}{*}{\textbf{Class}} & \multirow{2}{*}{\textbf{Real}} & \multicolumn{2}{c|}{\textbf{Weather}} & \multicolumn{2}{c|}{\textbf{Lighting}} 
            & \multicolumn{2}{c}{\textbf{Object}} \\ 
            % \cline{6-11}
            & & & & & & & & & \textbf{CL} & \textbf{RA} & \textbf{DT} & \textbf{NT} & \textbf{MS} & \textbf{MM} \\
            \hline
            N-Caltech ~\cite{ncaltech2015}                                 &2015   &Simulator         &-                  &9000 &9K &1-10s &101  
            &           &\checkmark &           &\checkmark &           &\checkmark &  \\
            SEVD ~\cite{aliminati2024sevdsyntheticeventbasedvision}        &2024   &Simulator         &-                  &- &9M &2-30m &6
            &           &\checkmark &\checkmark &\checkmark &\checkmark &\checkmark &  \\            
            DDD17 ~\cite{binas2017ddd17endtoenddavisdriving}               &2017   &DAVIS 346B        &346$\times$260px   &36 &- &1-50m &7
            &\checkmark &\checkmark &\checkmark &\checkmark &\checkmark &\checkmark &  \\
            % MVSEC~\cite{zhu2018mvsec}                                     &2018   &DAVIS 346B        &346$\times$260px   & & & &
            % &\checkmark &\checkmark &           &\checkmark &           &\checkmark &  \\
            DDD20 ~\cite{ddd202020}                                        &2020   &DAVIS 346B        &346$\times$260px   &216 &- &1-50m &7
            &\checkmark &\checkmark &           &\checkmark &\checkmark &\checkmark &  \\
            Gen1 ~\cite{detournemire2020gen1largescaleeventbaseddetection} &2020   &Prophesee Gen1    &304$\times$240px   &2357 &255K &30-120s &2  
            &\checkmark &\checkmark &           &\checkmark &           &\checkmark &  \\
            1Mpx ~\cite{perot20201mpxlearningdetectobjects1}               &2020   &Prophesee Gen2    &1280$\times$720px  &929 &25M &30-120s &3
            &\checkmark &\checkmark &           &\checkmark &           &\checkmark &  \\
            DSEC ~\cite{dsec2021}                                          &2021   &Prophesee Gen3.1  &640$\times$480px   &60 &390K &1-30m &8
            &\checkmark &\checkmark &           &\checkmark &\checkmark &           &  \\
            EvDET200K (Ours)  &2024   &Prophesee EVK4–HD &1280$\times$720px  &10054 &200K &2-5s &10 
            &\checkmark &\checkmark &\checkmark &\checkmark &\checkmark &\checkmark &\checkmark  \\
            \hline     
        \end{tabular}
        }
    \label{tab_bench_dataset}
\end{table*}

\begin{figure*}
\centering
\includegraphics[width=1\linewidth]{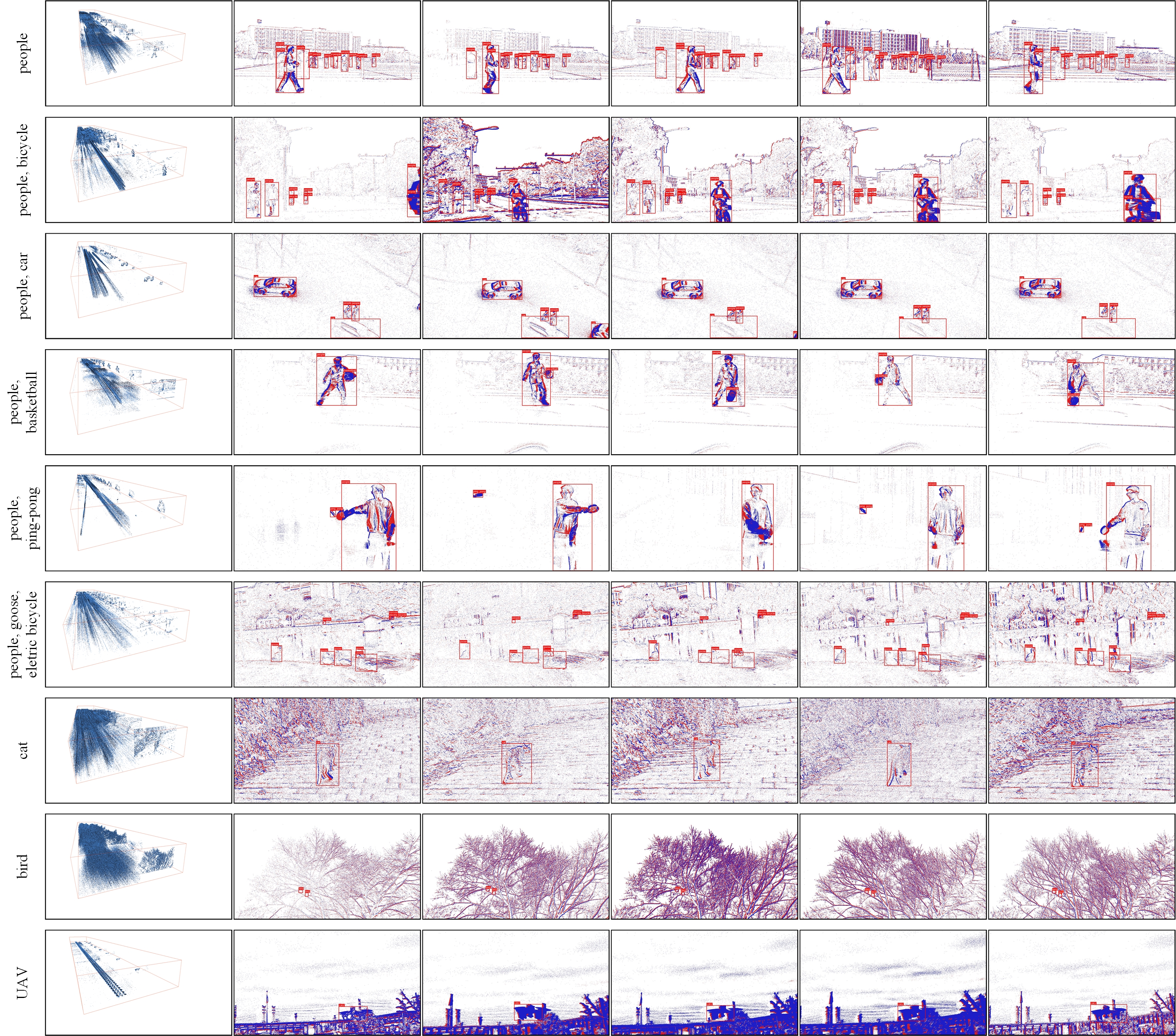}
\caption{ Illustration of some representative samples of our proposed EvDET200K dataset. 
}
\label{fig_dataset_visual}
\end{figure*}

\subsection{Data Collection and Annotation}
The EvDET200K dataset is captured using the PROPHESEE EVK4–HD event camera with a resolution of $1280 \times 720$. During the actual shooting process, we always adhere to the above principles to ensure that our proposed dataset contains rich event data and diverse challenges. 
% We will release our dataset in two formats: one is the raw event data stream, where each video contains all event information stored in CSV format as quadruples (x, y, p, t); the other format compresses the events into frames. 
% Event frames are widely used in current deep learning research. 
We convert each video into five frame images and manually annotate them. 
For the annotation, we use the XYXY format to store bounding-box coordinates, with each annotation represented as a five-tuple ($x_1$, $y_1$, $x_2$, $y_2$, $cls$), where $x_1$, $y_1$ denote the top-left corner of bounding-box and $x_2$, $y_2$ denote the bottom-right corner, along with the class label $cls$. The annotations for each video are saved in a JSON file.

\subsection{Statistical Analysis} 
\begin{figure}
\centering
\includegraphics[width=1\linewidth]{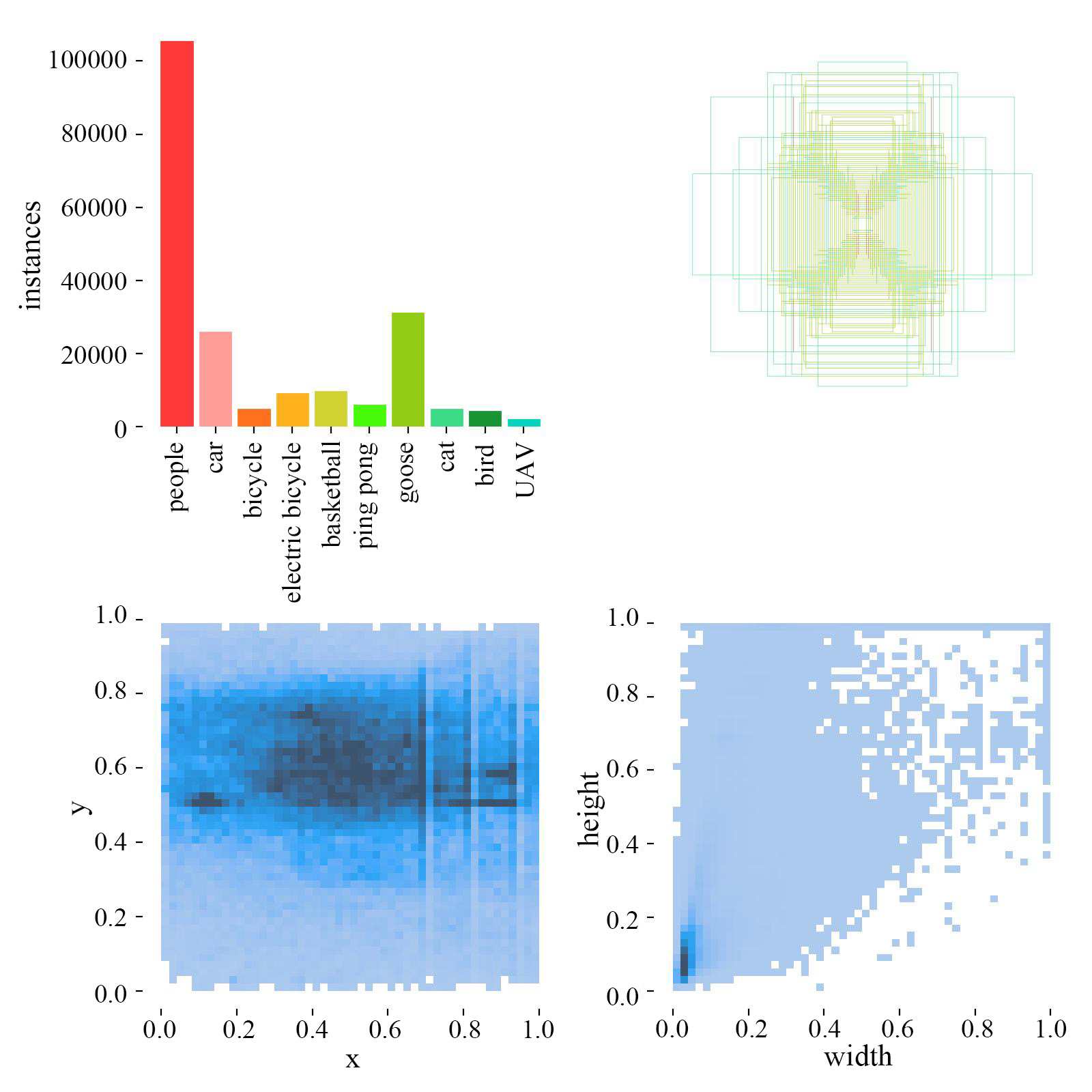}
\caption{\textbf{Visualization of All Annotation Information in the Dataset.} Top Left (Instance Count per Class) shows the number of instances for each class in the whole dataset; Top Right (Bounding Box Size Distribution)  illustrates the distribution of bounding box sizes across the dataset; Bottom Left (Object Center Distribution) shows the relative position (x, y coordinates) of object centers within the images. Bottom Right (Aspect Ratio Distribution) displays the distribution of width-to-height ratios of objects in the dataset.}
\label{fig_labels}
\end{figure}

The EvDET200K dataset comprises 10,054 video streams, annotated with 10 common object categories, totally 202,260 annotations. 
As shown in the upper left image of Fig.~\ref{fig_labels}, the most annotated class is ``people", with a total of 105,265 annotations. 
This is followed by ``goose" and ``car", which have 30,960 and 25,850 annotations, respectively. 
Among them, 2,949 videos are taken from dense scenes. Each video has a duration ranging from 2 to 5 seconds. 
The dataset is randomly divided into training, validation, and test subsets using the ratio 6:1:3, which contains 6,031, 1,002, and 3,021 video streams, respectively.

% \begin{figure}
% \centering
% \includegraphics[width=1\linewidth]{figures/dataset_bubble_chart.png}
% \caption{ An overview of the benchmarked event-based object detection dataset in this paper. }
% \label{fig_dataset_bubble_chart}
% \end{figure}

\subsection{Benchmark Baselines} 
To establish a comprehensive benchmark dataset for event-based object detection, we have selected more than 15 SOTA or representative detectors for evaluation on our proposed dataset, including: 
\textbf{1). CNN-based Detectors:} DetectoRS~\cite{detectors}, RED~\cite{perot20201mpxlearningdetectobjects1}, Mask R-CNN~\cite{he2018maskrcnn}, RetinaNet~\cite{lin2018focallossdenseobject}, and Faster R-CNN~\cite{ren2016fasterrcnnrealtimeobject}, all using ResNet-50 as the backbone, as well as YOLO-style detectors like YOLO v10~\cite{wang2024yolov10}, YOLO v6~\cite{li2022yolov6}.
\textbf{2). Transformer-based Detectors:} S5-ViT~\cite{zubić2024statespacemodelsevent}, SAST~\cite{peng2024sceneadaptivesparsetransformer}, RVT~\cite{gehrig2023recurrentvisiontransformersobject}, Swin-T~\cite{liu2021swintransformerhierarchicalvision}, DETR~\cite{carion2020endtoendobjectdetectiontransformers}, and vHeat~\cite{wang2024vheatbuildingvisionmodels}. 
\textbf{3). SNN-based Detectors:} Spiking neural network (SNN) detectors, such as spikeYOLO~\cite{luo2024integervaluedtrainingspikedriveninference} and EMS-YOLO~\cite{su2023deepdirectlytrainedspikingneural}, are also included for their event-based data processing capability.

% \begin{table}
%     \centering
%     \caption{Description of 10 categories in our EvDET200K dataset.}
%     \begin{tabular}{l|c}
%      \hline
%  ID &Attributes\\ \hline
%          01&people \\ 
%          02&car \\ 
%          03&bicycle \\ 
%          04&electric bicycle \\ 
%          05&basketball \\ 
%          06&ping pong \\ 
%          07&goose \\ 
%          08&cat \\ 
%          09&bird \\ 
%          10&UAV \\ \hline
%     \end{tabular}
    
%     \label{tab:my_label}
% \end{table}

\section{Experiments} 

% \subsection{Dataset and Evaluation Metric} 
% In addition to the newly proposed EvDET200K dataset, we also conducted a comparison with several state-of-the-art detectors on the N-Caltech101~\cite{ncaltech2015} dataset to validate the generalization capability of our method. 
% For evaluation metrics, we used the mean Average Precision (mAP) at different IoU thresholds, the most commonly used metric in object detection. 
% We also report Precision and Recall to assess the accuracy of predictions and the ability to detect positive instances. 
% Additionally, we measured the number of parameters, FLOPs, and FPS for each detector, providing a more comprehensive and accurate understanding of the models' performance.

\begin{table}[]
\caption{Experimental results on the N-Caltech101 dataset.}
\centering
\label{tab_res_ncal}
\begin{tabular}{l|l|c}
\hline
Methods & Format & mAP \\ \hline
NvS~\cite{li2023nvs} & Event Points & 34.6 \\
YOLE~\cite{yole} & Event Frames & 39.8 \\
Jeziorek et al.~\cite{Jeziorek_2024} & Event Frames & 53.4 \\
EAS-SNN~\cite{wang2024eassnn} & Event Points & 53.8 \\
\hline
Ours & Event Frames & 55.7 \\ \hline
\end{tabular}
\end{table}

\begin{table*}[!htp]
\centering
\caption{Experimental results on the newly proposed EvDET200K benchmark dataset. 
    vHeat$^{\dagger}$ means using vHeat as the encoder and Transformer as the decoder.}
    \label{tab:EvDET200Kresults}
\resizebox{\textwidth}{!}{ % 这里缩放表格以适应页面宽度
    \begin{tabular}{c|l|l|l|c|c|c|c|c|c|c|c|c}
    \hline 
    \textbf{Index} &\textbf{Algorithm} &\textbf{Publish}  &\textbf{Backbone}  &\textbf{mAP@50:95} &\textbf{mAP@50} &\textbf{mAP@75} &\textbf{P} &\textbf{R} &\textbf{Params} &\textbf{FLOPs} &\textbf{FPS} &\textbf{Code} \\
    \hline  
    01   &Faster R-CNN~\cite{ren2016fasterrcnnrealtimeobject}  &TPAMI 2016  &ResNet50  &46.0 &73.3 &48.6 &76.3 &88.3  &40.9M &71.2G  &23
    &\href{https://github.com/rbgirshick/py-faster-rcnn}{URL} \\      
    \hline     
    02   &S5-ViT~\cite{zubić2024statespacemodelsevent}  &CVPR 2024  &Former+SSM  &42.9 &76.3 &44.1 &69.7 &66.9 &18.2M &5.6G  &84 
    &\href{https://github.com/uzh-rpg/ssms_event_cameras}{URL} \\ 
    \hline  
    03   &SAST~\cite{peng2024sceneadaptivesparsetransformer}  &CVPR 2024  &Transformer  &27.4  &53.6 &25.3 &30.1 &29.6 &18.5M &15.9G  &51 &\href{https://github.com/Peterande/SAST}{URL} \\
    \hline  
    04   &SpikeYOLO~\cite{luo2024integervaluedtrainingspikedriveninference}  &ECCV 2024  &SNN  &41.2  &74.8 &39.8 &\textbf{81.6} &68.5  &68.8M &78.1G  &77  &\href{https://github.com/BICLab/SpikeYOLO}{URL} \\
    \hline      
    \multirow{4}{*}{05} & YOLOv10-N~\cite{wang2024yolov10} & \multirow{4}{*}{arXiv 2024} &\multirow{4}{*}{CNN}  &42.7 &75.1 &42.1 &75.3 &68.9 &2.3M &8.2G &116 &\multirow{4}{*}{\href{https://github.com/THU-MIG/yolov10}{URL} }\\
    &YOLOv10-S~\cite{wang2024yolov10} & & &43.5 &76.2 &43.0 &75.4 &71.2 &7.3M  &21.6G &83 &\\
    &YOLOv10-M~\cite{wang2024yolov10} & & &44.0 &77.5 &42.8 &76.6 &71.7 &15.4M &59.1G &32  &\\
    &YOLOv10-B~\cite{wang2024yolov10} & & &44.1 &77.9 &43.1 &76.0 &73.2 &19.1M  &92.0G &30 &\\
    \hline  
    06   &RVT~\cite{gehrig2023recurrentvisiontransformersobject}  &CVPR 2023  &Transformer  &40.7 &73.1 &42.3 &70.3 &65.9 & 9.9M &8.4G &88 
    &\href{https://github.com/uzh-rpg/RVT}{URL} \\
    \hline  
    07   &EMS-YOLO~\cite{su2023deepdirectlytrainedspikingneural}  &ICCV 2023  &SNN  &32.1  &66.6 &27.4 &77.5 &62.5 &14.40M &3.3M  &119 
    &\href{https://github.com/BICLab/EMS-YOLO}{URL} \\
    \hline  
    08   &YOLOv6~\cite{li2022yolov6}  &arXiv 2022  &RepVGG  &41.3  &75.7 &38.9 &50.4 &53.8 &17.2M &44.2M  &70
    &\href{https://github.com/meituan/YOLOv6}{URL} \\
    \hline  
    09   &Swin-T~\cite{liu2021swintransformerhierarchicalvision}  &ICCV 2021  &Transformer  &49.0 &78.4 &52.7 &\textbf{79.4} &88.8  &160M &1043G  &26 &\href{https://github.com/microsoft/Swin-Transformer}{URL} \\
    \hline  
    10   &DetectoRS~\cite{detectors}  &CVPR 2021  &ResNet50  &49.1 &78.8 &53.5 &78.8 &85.8   &123.2M &117.2G  &32
    &\href{https://github.com/joe-siyuan-qiao/DetectoRS}{URL} \\
    \hline  
    11   &RED~\cite{perot20201mpxlearningdetectobjects1} &NeurIPS 2020  &ResNet50  &35.4 &68.3 &35.2 &69.0 &66.1  &24.1M &46.3G  &34
    &\href{https://github.com/prophesee-ai/prophesee-automotive-dataset-toolbox}{URL} \\
    \hline  
    12   &DETR~\cite{carion2020endtoendobjectdetectiontransformers}  &ECCV 2020  &Transformer  &40.9 &74.5 &39.5 &74.5 &\textbf{90.9}  &41M &86G  &29 &\href{https://github.com/facebookresearch/detr}{URL} \\
    \hline  
    13   &Mask R-CNN~\cite{he2018maskrcnn}  &ICCV 2017  &ResNet50  &48.8 &77.6 &52.0 &77.8 &87.1  &43.8M &142.7G  &28
    &\href{https://github.com/facebookresearch/Detectron}{URL} \\
    \hline  
    14   &RetinaNet~\cite{lin2018focallossdenseobject}  &ICCV 2017  &ResNet50  &48.6 &77.0 &50.8 &77.8 &\textbf{93.7}  &36.2M &81.4G  &76 &\href{https://github.com/facebookresearch/Detectron}{URL} \\
    \hline
    15   &vHeat$^{\dagger}$~\cite{wang2024vheatbuildingvisionmodels}  &arXiv 2024  &vHeat  &50.3 &72.2 &54.2 &56.9 &69.4 &56.3M &74.5G &50 
    &\href{https://github.com/uzh-rpg/ssms_event_cameras}{URL} \\    
    \hline      
    16   &Ours  &-  &MvHeat  &\textbf{52.9}      &\textbf{80.4}       &\textbf{55.9}       &58.9       &70.0       &47.5M       &56.4G       &58    &-   \\ 
    \hline     
    \end{tabular}
    }
\end{table*}

% \subsection{Implementation Details} 
% For training the detector, we set the number of epochs to 80. The model is optimized using the AdamW optimizer with an initial learning rate of 0.001 and weight decay of 0.0001. The batch size is set to 6, and the input image size is 640x640. Our code is implemented in Python using the PyTorch framework, and the experiments are conducted on a server equipped with an AMD EPYC 7542 32-Core Processor CPU and an NVIDIA RTX 4090 GPU. This configuration ensures efficient training and helps the model achieve stable convergence through the use of the AdamW optimizer's adaptive learning rate and regularization.

\subsection{Compare With Other Detectors}

\textbf{Results on EvDET200K Dataset.}
% As shown in Tab.~\ref{tab:EvDET200Kresults}, our baseline $vHeat^{\dagger}$ achieves 50.3/72.2/54.2 on mAP/mAP@50/mAP@75, meanwhile, our model EvHeat-DET achieves 52.9/80.4/55.9, which is significantly better than baseline. Obviously, our detector also better than other SOTA detectors including R-CNN based methods (like Faster R-CNN, Mask R-CNN), YOLO-based (YOLOv10, YOLOv6) detectors , SNN-based detectors (EMS-YOLO, SpikeYOLO) and Transformer-based methods (DETR, RVT, SAST, S5-ViT). This result strongly demonstrates the effectiveness of our method. We also caculate model parameter, FLOPs and FPS to comprehensively summarize the characteristics of each model.
As shown in Tab.~\ref{tab:EvDET200Kresults}, our baseline $vHeat^{\dagger}$ achieves 50.3/72.2/54.2 on mAP/mAP@50/mAP@75, meanwhile, our model EvHeat-DET achieves 52.9/80.4/55.9, which is significantly better than baseline. Obviously, our detector also better than other SOTA detectors including R-CNN based methods, YOLO-based detectors, SNN-based detectors and Transformer-based methods. This result strongly demonstrates the effectiveness of our method.

\noindent \textbf{Results on N-Caltech Dataset.}
% Tab.~\ref{tab_res_ncal} presents experimental results on the N-Caltech101 dataset, comparing methods in terms of mean Average Precision (mAP). Among the methods listed, NvS (using event points) achieves a mAP of 34.6, while YOLE and Jeziorek et al. (both using event frames) achieve mAPs of 39.8 and 53.4, respectively. The EAS-SNN method, which also uses event points, achieves a slightly higher mAP of 53.8. Our model MvHeat-DET outperforms all others with a mAP of 55.7 using event frames.
Tab.~\ref{tab_res_ncal} presents experimental results on the N-Caltech101 dataset, comparing methods in terms of mean Average Precision (mAP). Among the methods listed, NvS and EAS-SNN (using event points) achieve 34.6/53.8, while YOLE and Jeziorek et al. (both using event frames) achieve mAPs of 39.8 and 53.4, respectively. Our model MvHeat-DET outperforms all others with a mAP of 55.7.

\subsection{Component Analysis} 
We use DETR as the base model for component analysis. IQS denotes adding an IoU-based query selection strategy to the base model, vEnc. indicates replacing the transformer encoder with vHeat, and MoE represents using a multi-expert strategy. Tab.~\ref{tab_abl_component} shows that the base model achieves 40.9 mAP on the EvDET200K dataset. Adding IQS to optimize query selection raises the detection result to 41.6. Next, replacing DETR's encoder with vHeat significantly improves accuracy, reaching 50.3. Finally, the introduction of a multi-expert mechanism further boosts the model to 52.9. With each additional component, the mAP steadily increases, indicating that each component contributes to model's performance. The notable improvements from IQS and vHeat suggest that these components significantly enhance feature extraction and information processing capabilities. The addition of MoE also further refines the model’s performance.

\begin{table}[]
\centering
\caption{Component Analysis on Our Proposed EvDET200K dataset. IQS means add IoU-based Query selection strategy, vEnc. means replace the encoder with vHeat Encoder, MoE means add MoE strategy to encoder.}
\label{tab_abl_component}
    \begin{tabular}{c|cccc|c}
    \hline
        % \multirow{2}{*}{\textbf{No.}} & \multicolumn{4}{c|}{\textbf{Component}} & \multirow{2}{*}{\textbf{mAP}} \\
        %  & \textbf{Base.} & \textbf{IQS} & \textbf{vEnc.} & \textbf{MoE} &  \\ \cline{1-6}
        \textbf{Index} &\textbf{Baseline}  &\textbf{IQS}  &\textbf{vEnc}.  &\textbf{MoE}  & \textbf{mAP} \\
        \hline 
        1 &\checkmark  &  &  &  & 40.9 \\
        2 &\checkmark  &\checkmark  &  &  & 41.6 \\
        3 &\checkmark  &\checkmark  &\checkmark &  & 50.3 \\
        4 &\checkmark  &\checkmark  &\checkmark  &\checkmark  & 52.9 \\ 
    \hline
    \end{tabular}
\end{table}

\subsection{Ablation Study}  

% $\bullet$ \noindent \textbf{Analysis on Different Resolutions of Event Stream.~}  
% $\bullet$ \noindent \textbf{Analysis on Number of MHCO in each Stage.~}   
% $\bullet$ \noindent \textbf{Analysis on the channel of MHCO in different stage.~}   
% $\bullet$ \noindent \textbf{Analysis on the number of expert.~}   
% $\bullet$ \noindent \textbf{Analysis on thermal diffusivity \(k\).~}  

\textbf{Analysis on Different Resolutions of Event Stream.} 
% In this section, we investigate the impact of event stream resolution on detection performance. We conducted experiments with four different resolutions: 256$\times$256px, 448$\times$448px, 512$\times$512px, and 640$\times$640px. As shown in Tab.~\ref{tab_abl_res}, we observe that the 256$\times$256px resolution achieves 40.0/69.7/39.5, 448$\times$448px achieves 49.7/78.4/52.1, 512$\times$512px achieves 52.0/79.9/54.7, and 640$\times$640px achieves 52.9/80.4/55.8. Intuitively, higher resolution event streams retain more spatial information, which can positively influence the model's performance. However, higher resolution also demands more computational power and time, resulting in increased FLOPs. Our model's performance at lower resolutions is not particularly outstanding, but it still achieves strong results at intermediate resolutions, demonstrating its adaptability to scenarios with limited computational resources.
In this section, we investigate the impact of event stream resolution on detection performance. We conduct experiments with four different resolutions: 256$\times$256px, 448$\times$448px, 512$\times$512px, and 640$\times$640px. As shown in Tab.~\ref{tab_abl_res}, we observe that the 256$\times$256px resolution achieves 40.0/69.7/39.5, 448$\times$448px achieves 49.7/78.4/52.1, 512$\times$512px achieves 52.0/79.9/54.7, and 640$\times$640px achieves 52.9/80.4/55.8. Intuitively, higher resolution event streams retain more spatial information, which can positively influence the model's performance. Our model's performance at lower resolutions is not particularly outstanding, but it still achieves strong results at intermediate resolutions, demonstrating its adaptability to scenarios with limited computational resources.

\begin{table}
\centering
\caption{Ablation studies on different input resolution. }
\resizebox{\linewidth}{!}{
\begin{tabular}{l|cccc}
\hline
\textbf{Resolution}& \textbf{mAP} &\textbf{mAP@50} &\textbf{mAP@75} &\textbf{FLOPs}\\
\hline
256$\times$256px& 40.0 &69.7 &39.5 &11.9G\\
448$\times$448px& 49.7 &78.4 &52.1 &36.3G \\
512$\times$512px& 52.0 &79.9 &54.7 &47.2G \\
640$\times$640px& 52.9 &80.4 &55.8 &73.4G \\
\hline
\end{tabular}
}
\label{tab_abl_res}
\end{table}

\noindent \textbf{Analysis on Number of MHCO in Each Stage. }
Tab.~\ref{tab_abl_MHCO} shows the effect of varying the number of MHCO modules on model performance, computational complexity, and parameter count. We set the input event stream resolution to 640$\times$640px and varied the number of MHCO modules in the third stage. Specifically, increasing the number of MHCO modules had a positive impact on the experimental results (6 layers: 52.6, 12 layers: 52.9, 18 layers: 52.9, 24 layers: 53.4). While increasing the number of MHCO modules can improve model accuracy, both FLOPs and the number of parameters grow significantly, leading to a substantial rise in computational and storage demands. We choose the (2, 2, 12, 2) configuration for building the experimental model to achieve a balance between model accuracy and computational efficiency while managing resource consumption.
\begin{table}
    \centering
    \caption{Ablation studies on Number of MHCO in Each Stage. }
    \resizebox{\linewidth}{!}{
    \begin{tabular}{l|cccc}
        \hline
         \textbf{Number of MHCO}& (2,2,6,2) &(2,2,12,2) &(2,2,18,2) &(2,2,24,2)\\
         \hline
         \textbf{mAP}           & 52.6      &52.9       &52.9       &53.4 \\
         \textbf{FLOPs}         & 39.3G     &56.4G      &73.4G      &90.5G \\
         \textbf{Param}         & 36.2M     &47.5M      &65.1M      &70.1MG \\
        \hline
    \end{tabular}
    }
    \label{tab_abl_MHCO}
\end{table}

\noindent \textbf{Analysis on Different Channels in Each Stage.}
Tab.~\ref{tab_abl_channel} presents the changes in mAP, FLOPs, and parameters at different stages of the model with different channel configurations. We list two MHCO configurations: (2,2,6,2) and (2,2,18,2), each paired with two sets of channel configurations: (64, 128, 256, 512) and (96, 192, 384, 768). Specifically, with the (2,2,6,2) configuration, mAP increases from 50.4 to 52.6; with the (2,2,18,2) configuration, mAP rises from 51.9 to 52.9. This indicates that increasing the number of channels contributes to performance improvement. However, increasing the number of channels also leads to higher computational costs and an increase in the number of parameters. For the (2,2,6,2) configuration, FLOPs increase from 18.0G to 39.2G, and the number of parameters increases from 19.4M to 36.2M. It is important to balance the performance gain with the computational and storage overhead, depending on the specific application and hardware constraints.
% Tab.~\ref{tab_abl_channel} presents the changes in mAP, FLOPs, and the number of parameters at different stages of the model with different channel configurations. We list two MHCO configurations, (2,2,6,2) and (2,2,18,2), each paired with two sets of channel configurations: (64, 128, 256, 512) and (96, 192, 384, 768). The results show that increasing the number of channels (from (64,128,256,512) to (96,192,384,768)) generally improves model accuracy. Specifically, with the (2,2,6,2) configuration, mAP increases from 50.4 to 52.6; with the (2,2,18,2) configuration, mAP rises from 51.9 to 52.9. This indicates that increasing the number of channels contributes to performance improvements. However, increasing the number of channels also leads to higher computational costs and an increase in the number of parameters. With the (2,2,6,2) configuration, FLOPs increase from 18.0G to 39.2G, and the number of parameters increases from 19.4M to 36.2M. With the (2,2,18,2) configuration, FLOPs grow from 33.1G to 73.4G, and parameters increase from 29.7M to 58.7M. Thus, while increasing the number of channels improves model accuracy, it also significantly increases computational complexity and storage requirements. Therefore, when choosing the number of channels, it is important to balance the performance gain with the computational and storage overhead, depending on the specific application and hardware constraints.

\begin{table}[]
    \centering
    \caption{Ablation studies on different channels in each stage.}
    \label{tab_abl_channel}
    \resizebox{\linewidth}{!}{
    \begin{tabular}{c|cc|cc}
    \hline
        \textbf{Layer}   & \multicolumn{2}{c|}{(2,2,6,2)} & \multicolumn{2}{c}{(2,2,18,2)} \\ \cline{2-5} 
        \textbf{Channel} & \multicolumn{1}{c|}{(64,128,256,512)} & (96,192,384,768) & \multicolumn{1}{c|}{(64,128,256,512)} & (96,192,384,768) \\ \hline
        \textbf{mAP}     &50.4  & 52.6 &51.9  & 52.9 \\
        \textbf{FLOPs}   & 18.0G & 39.2G & 33.1G & 73.4G \\
        \textbf{Param}   & 19.4M & 36.2M & 29.7M & 58.7M \\ \hline
    \end{tabular}
    }
\end{table}

\begin{figure*}[!htp]
\centering
\includegraphics[width=1\linewidth]{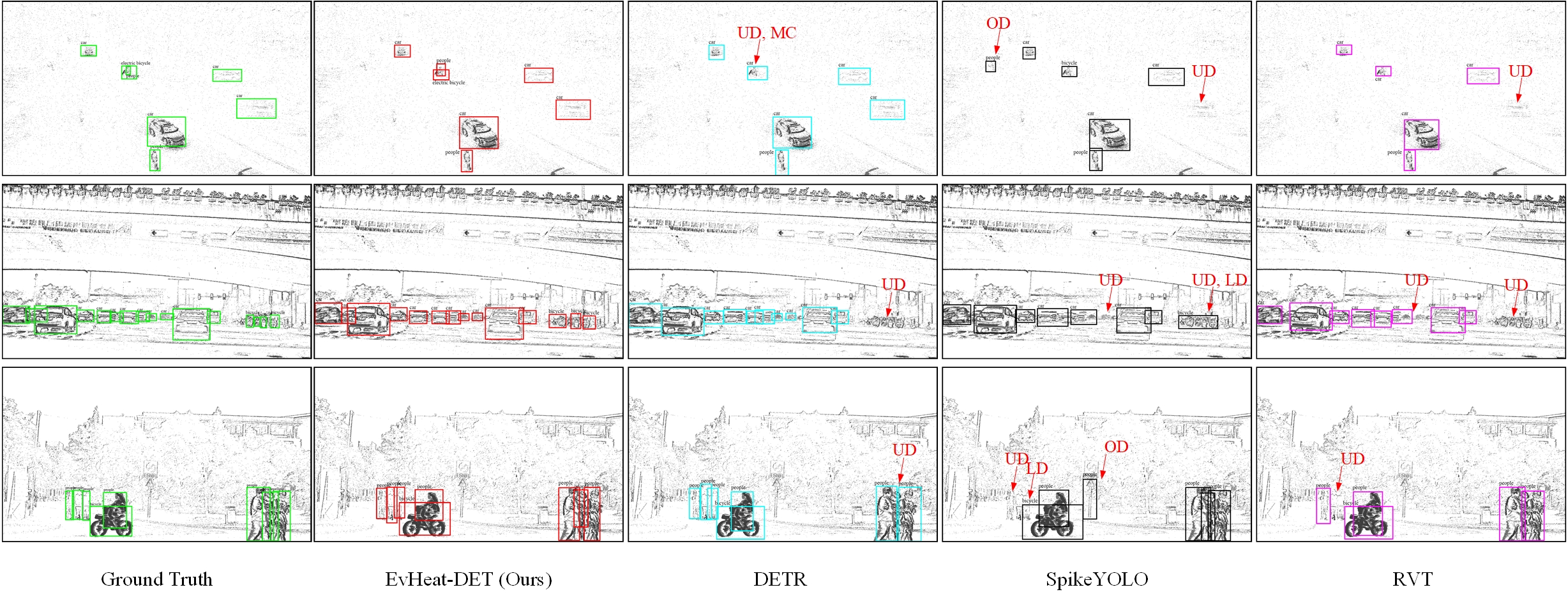}
\caption{Visualization of the detection results of ours and other detectors. (MC: misclassification, UD: undetected, OD: over-detected, LD: large deviation.)}
\label{fig_det_res}
\end{figure*}

\begin{figure*}[!htp]
\centering
\includegraphics[width=1\linewidth]{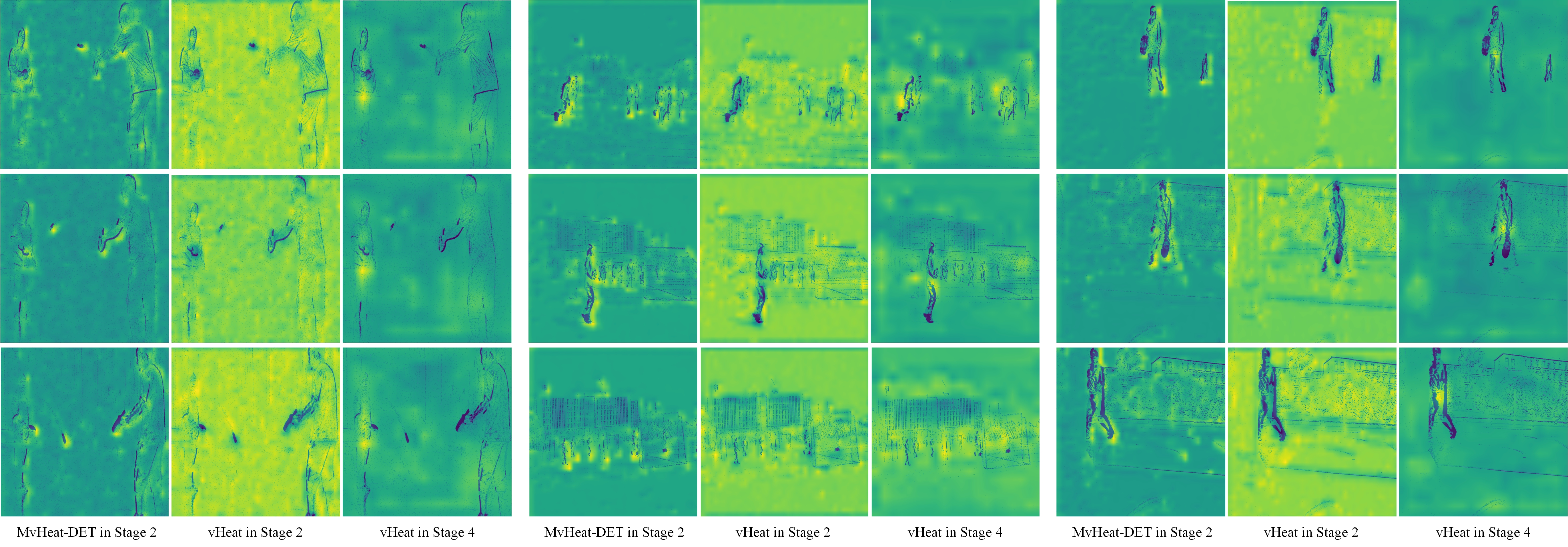}
\caption{Visualization of the feature maps compared with vHeat.}
\label{fig_feat_map_compare}
\end{figure*}

\noindent \textbf{Analysis on Number of Expert.}
In this section, we investigate the impact of varying the number of experts on the experimental results. As shown in Tab.~\ref{tab_abl_expert}, we select three transformations (DCT, DFT, and HT) as the experts. As the number of experts increases, the mAP gradually improves, rising from 50.3 with one expert (DCT) to 52.7 with two experts (DCT + DFT), and further to 52.9 with three experts (DCT + DFT + HT). This indicates that increasing the number of experts helps enhance model performance, primarily because different types of experts provide diverse feature extraction capabilities. It also demonstrates the effectiveness of the proposed MHCO module.
\begin{table}
    \centering
    \caption{Ablation studies on the number of expert.}
    \begin{tabular}{lc}
        \hline
         \textbf{Number of Expert}& \textbf{mAP}\\
         \hline
         1 (DCT)         & 50.3\\
         2 (DCT+DFT)     & 52.7 \\
         3 (DCT+DFT+HT)  & 52.9\\
        \hline
    \end{tabular}
    \label{tab_abl_expert}
\end{table}

\noindent \textbf{Analysis on thermal diffusivity \(k\).}
The Tab.~\ref{tab_abl_k} presents experimental results on different settings for the thermal diffusivity \(k\), evaluating its impact on model performance (mAP). When \(k\) is fixed, the model achieves the mAP of 48.2. When \(k\) is treated as a learnable parameter, the mAP increases to 49.1, indicating that allowing the model to automatically learn \(k\) improves performance to some extent. When \(k\) is predicted using FEs, the mAP further increases to 49.7, achieving the best performance. Dynamic learning and prediction of thermal diffusivity \(k\), particularly through frequency embedding, significantly improve the model's accuracy and allow it to better adapt to frequency feature variations in the input data.
% These results suggest that dynamic learning and prediction of thermal diffusivity \(k\), particularly through frequency embedding, significantly improve the model's accuracy and allow it to better adapt to frequency feature variations in the input data.

\begin{table}[]
\caption{Ablation studies on thermal diffusivity \(k\). }
\label{tab_abl_k}
\centering
    \begin{tabular}{ll}
    
        \hline
        \textbf{Settings}            & \textbf{mAP} \\ \hline
        Fixed \(k\)                  & 48.2 \\
        \(k\) as learnable parameter & 49.1 \\
        Predicting \(k\) using FEs   & 49.7 \\ \hline
    \end{tabular}
\end{table}
% \subsection{Efficiency Analysis} 

\section{Visualization}  

$\bullet$ \noindent \textbf{Detection Results.~}  
The comparison shown in Fig.~\ref{fig_det_res} illustrates the detection results of our proposed MvHeat-DET alongside DERT, SpikeYOLO, and RVT detectors. As seen in the figure, our detector demonstrates strong performance even in dense scenes, whereas the other detectors tend to suffer from missed detections or false positives in such environments.

\noindent $\bullet$ \noindent \textbf{Feature Maps.~}  
Fig.~\ref{fig_feat_map_compare} shows representative feature map visualizations of our proposed method on the EvDET200K dataset. It is evident that, even in challenging scenarios, our method is still able to focus on the key detection areas, demonstrating the effectiveness of our model. It also compares the feature maps of our method with the vHeat model. Our model is able to focus on the detection targets effectively from the second stage, while vHeat only begins to focus on the key areas at the fourth stage. This indicates that we can design smaller network architectures to achieve a better balance between performance and calculation consumption.

% \section{Limitation Analysis}  
% There is still room for improving our model. For instance, when the model is able to obtain good feature outputs in the shallow layers (as shown in Fig.~\ref{fig_feat_map_compare}), we can consider reducing the number of layers to decrease the model's complexity and achieve higher detection efficiency. Additionally, the current model does not fully leverage the temporal information. In the future, we will explore the use of 3D heat conduction inference, combined with the rich temporal information in the event stream, to enable more efficient detection.

\section{Conclusion}  
In this paper, we introduce a novel approach to event stream-based object detection, termed MvHeat-DET, which leverages a MoE-based heat conduction framework for efficient and interpretable feature extraction. This method balances performance, efficiency, and interoperability, offering a promising solution to the challenges faced by existing event-based detectors. Additionally, we propose a new high-definition dataset, EvDET200K, designed to advance research in this field by providing a comprehensive benchmark with diverse object categories and samples. We conclude by re-training more than 15 state-of-the-art object detectors on this dataset, paving the way for future advancements in event-based object detection.

{
\small
\bibliographystyle{ieeenat_fullname}
\bibliography{main}
}

% WARNING: do not forget to delete the supplementary pages from your submission 
% \input{sec/X_suppl}

\end{document}